\documentclass[letterpaper, 10 pt, conference]{ieeeconf}  

\IEEEoverridecommandlockouts                              

\usepackage{graphicx}
\usepackage[colorinlistoftodos]{todonotes}
\usepackage{acronym}
\usepackage{adjustbox}
\usepackage{subfigure}
\usepackage{color}

\title{\LARGE \bf
Human-Robot Collaboration System Setup for Weed Harvesting Scenarios in Aquatic Lakes
}

\author{Ahmed H. Elsayed$^{1}$, Andrej Lejman$^{1}$ and Frederic Stahl$^{1}$
\thanks{$^{1}$German Research Center for Artificial Intelligence
GmbH (DFKI), Lower Saxony, Oldenburg, 26129, Germany
        {\tt\small ahmed.elsayed@dfki.de}%
}}

\newacro{SSS}{Side Scan Sonar}
\newacro{SONAR}{SOund NAvigation and Ranging}
\newacro{AWBs}{Artificial Water Bodies}
\newacro{USV}{Unmanned Surface Vehicle}

\usepackage{varioref} 
\usepackage[hidelinks]{hyperref}
\usepackage[noabbrev]{cleveref}
\usepackage{siunitx}  
\usepackage{tabularx}

\DeclareSIUnit{\kn}{knots}

\begin{document}

\maketitle
\thispagestyle{empty}
\pagestyle{empty}

\begin{abstract}
\ac{AWBs} are human-made and require continuous monitoring due to their artificial biological processes. These systems necessitate regular maintenance to manage their ecosystems effectively. \ac{USV} offers a collaborative approach for monitoring these environments, working alongside human operators such as boat skippers to identify specific locations. This paper discusses a weed harvesting scenario, demonstrating how human-robot collaboration can be achieved, supported by preliminary results. The \ac{USV} mainly utilises multibeam \ac{SONAR} for underwater weed monitoring, showing promising outcomes in these scenarios.
\end{abstract}

\section{INTRODUCTION}

\ac{AWBs} are created by humans for different reasons such as water retention in dam construction, urban development, rainwater storage, or leisure activities. Unlike natural lakes, \ac{AWBs} lack an established ecosystem, making them prone to environmental issues like ecological degradation and the spread of diseases if not properly monitored and maintained \cite{hunter1982man}. Managing \ac{AWBs}, including artificial lakes, presents unique challenges due to their reliance on continuous human intervention, given the unnatural balance of their ecosystems and the potential for disturbances in biological processes \cite{cantonati2020characteristics}.

This paper focuses on Maschsee Lake in Hannover, Germany; an artificial lake where rapid weed growth poses a significant problem, requiring continuous harvesting. Effective monitoring could reduce human labour and improve the efficiency of maintaining the lake's ecosystem. The proliferation of weeds negatively impacts the lake's biological life, affecting fish and other aquatic species, and disrupting leisure activities such as kayaking. Weeds can obstruct the propellers of small boats, posing risks to humans and equipment. 

A heterogeneous approach involving collaboration between human operators and \ac{USV}s offers a promising solution for managing the lake environment. The \ac{USV} can autonomously survey designated areas of the lake, providing real-time data to the boat skipper to facilitate advanced planning. This approach aims to reduce the skipper's workload, increase operational efficiency, and contribute to energy conservation and environmental sustainability.

While optical cameras have been employed for monitoring of aquatic vegetation \cite{gerlo2023seaweed}, \ac{SONAR} and backscatter strength data are often more effective for underwater vegetation detection \cite{mutlu2023acoustic, komatsu2003use}. \ac{SONAR} is particularly advantageous due to its ability to scan larger areas at greater depths compared to optical cameras, though it requires specialised expertise for operation and data interpretation. By leveraging \ac{SONAR}, we can achieve more efficient monitoring of underwater vegetation.

\section{METHODOLOGY \& MISSION SETUP}

\begin{figure}[t]
  \centering
  \includegraphics[width=0.7\columnwidth, angle=-90]{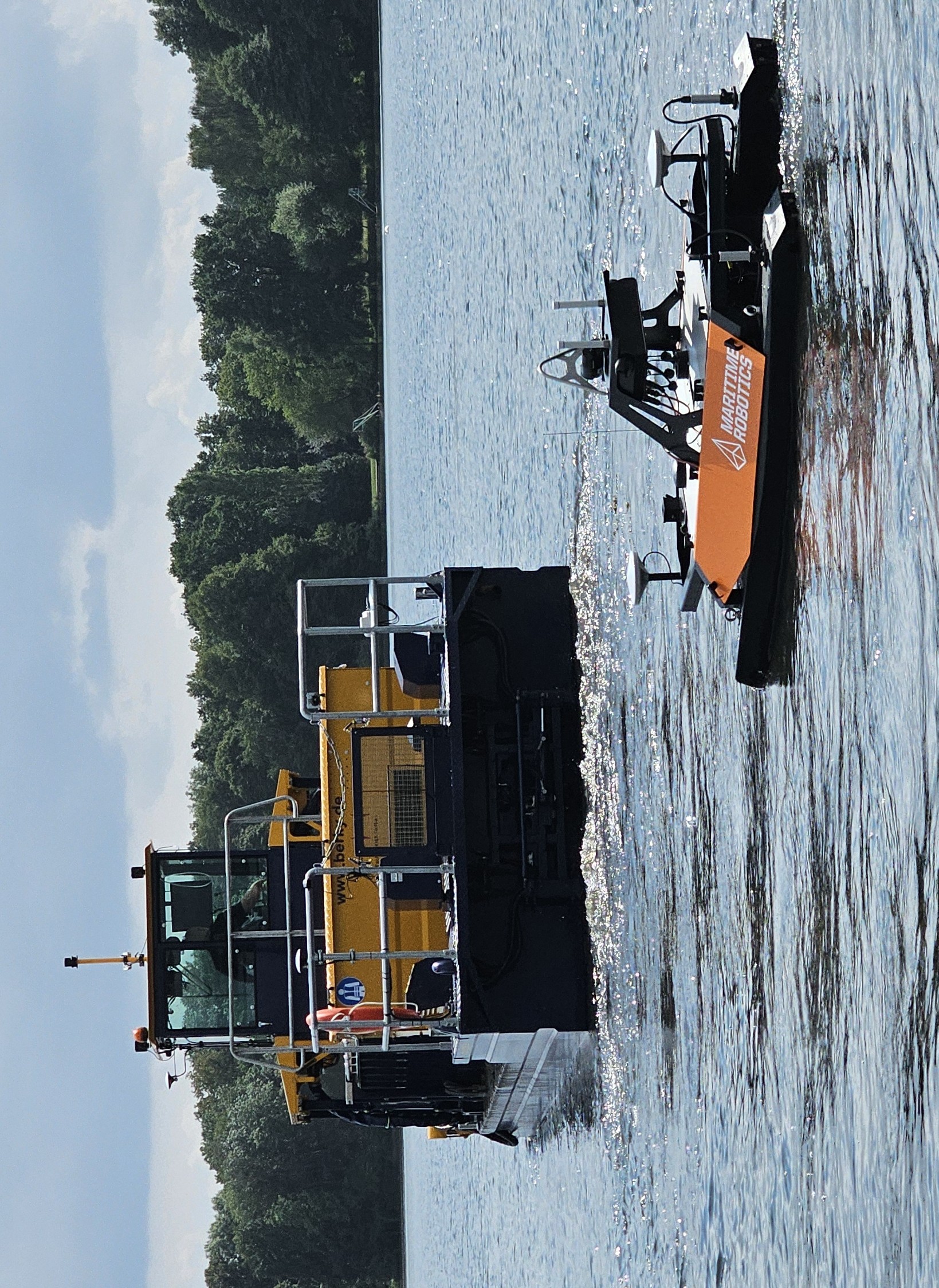}
  \caption{Weed Harvester boat and \ac{USV} collaborating in Maschsee Lake.}
  \label{fig:otter_berky}
\end{figure}

The mission aims to support the weed harvesting process in Maschsee Lake by adopting a human-robot collaboration approach. This collaboration combines the advantages of \ac{USV}s with human operators, to identify and map underwater weed clusters as depicted in figure \ref{fig:otter_berky}. The \ac{USV} performs a comprehensive survey of the lake using multibeam \ac{SONAR}, resulting in high-resolution maps that provide boat skippers with detailed information on the location and extent of weed growth. This approach allows for more energy-efficient and targeted harvesting operations, improving both operational efficiency and environmental management.

Before harvesting, a preliminary survey of the lake is performed using the iWBMS multibeam \ac{SONAR} from Norbit,\footnote{https://norbit.com/subsea/products, accessed on 13-09-2024} which provides a comprehensive scan of the underwater environment and  high-resolution bathymetry. This process generates detailed maps, providing insights into the distribution and density of weed growth. The \ac{SONAR} generates a bathymetric map, enabling accurate measurement of depth and the mapping of weed distribution throughout the lake. Additionally, backscatter data analysis allows for distinguishing between different underwater targets, such as weed clusters, the seabed, and other objects.

The multibeam \ac{SONAR} operates at a mean frequency of \SI{400}{\kilo\hertz}, with an \SI{80}{\kilo\hertz} wide chirp forming 256 beams, each with a beam resolution of \SI{0.9}{\degree}. The upper gate is set to \SI{1.0}{\metre}, and the lower gate to \SI{5.0}{\metre}, with a swath of \SI{150}{\degree}. These parameters allow for a detailed and accurate mapping of the underwater environment. The average survey speed was \SI{3}{\kn}.

\begin{figure}[htp]
  \centering
  \includegraphics[width=\columnwidth]{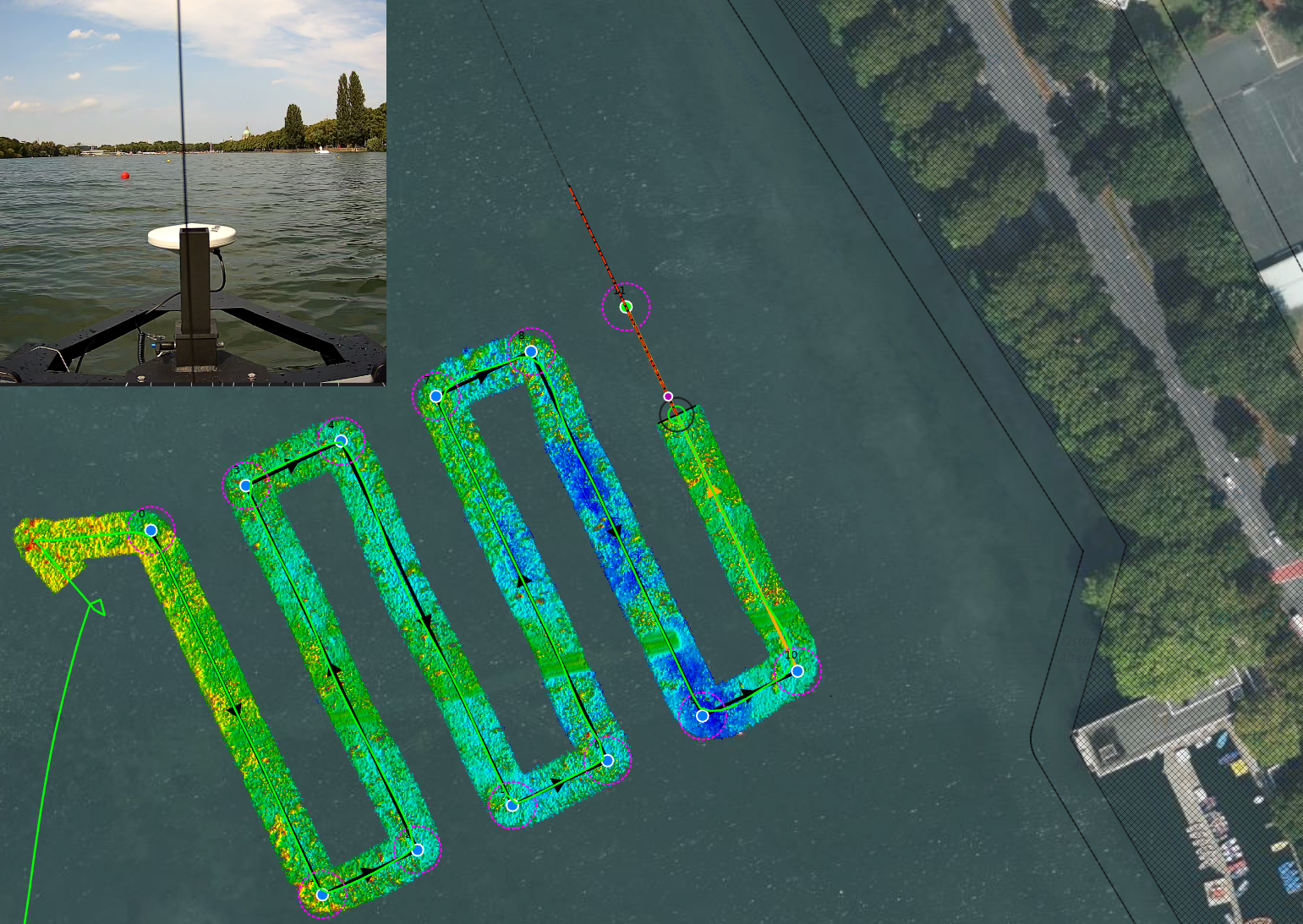}
  \caption{Georeferenced map of Maschsee Lake with RGB image captured by the \ac{USV}'s camera overlaid. The \ac{SONAR} data, visualising detected underwater features, is displayed along the predefined path of the \ac{USV}. The map shows the surveyed areas and the nature of the detected features - whether underwater vegetation, the seabed, or other objects - requiring further analysis}
  \label{fig:vcs_rgb}
\end{figure}

The \ac{USV} communicates directly through the interface used by the boat skippers. The map generated by the \ac{USV} is updated in real-time, allowing the skippers to visualise weed locations on their displays. The interface bridges the communication between the \ac{USV} and the skippers. As the \ac{USV} performs the survey, it provides continuous updates to the map with the latest location of weed clusters. This real-time feedback allows the boat skippers to plan their harvesting routes efficiently. The \ac{USV} transmits georeferenced acoustic images to the interface, which are overlaid with the lakes' topography to create a clear visual representation of weed distribution as shown in \cref{fig:vcs_rgb}.

Figure \ref{fig:otter_berky} shows the heterogeneous mission setup, in which the \ac{USV} follows the weed harvester boat. The mission is structured so that the \ac{USV} first scans a defined area using its multibeam \ac{SONAR} to map the underwater weed clusters before harvesting begins. Once the area has been fully scanned, the weed harvester boat follows the path of the \ac{USV}, mowing and collecting the weeds. To evaluate the impact of the harvesting process, the \ac{USV} scans the same area again after the harvester passes, enabling a comparison of the pre-harvest and post-harvest conditions of the lake.

In addition to \ac{SONAR} data, visual images captured by a GoPro camera attached to the \ac{USV} provide a more comprehensive representation of the underwater environment. These images provide additional information than the acoustic data, improving the visibility of areas where \ac{SONAR} data may be insufficient for precise interpretation. 

Furthermore, an oceanographic instrument enhances the mission by collecting essential environmental data. The AML-3 LGR sensor, manufactured by AML Oceanographic,\footnote{https://amloceanographic.com/aml-3-flexible-oceanographic-instrument, accessed on 13-09-2024} is used for precise water column profiling during the mission. Equipped with a lithium-ion battery and on-board memory, the sensor stores collected data, which is transferred via WiFi or USB. The AML-3 sensor provides data for multibeam \ac{SONAR} correction, including sound velocity and pressure profiling, ensuring accurate underwater mapping. It also monitors environmental conditions such as turbidity and dissolved oxygen levels, which influence weed growth. This additional information provides an understanding of the lake's ecosystem, providing valuable insights for monitoring and maintaining the artificial lake.

\section{PRIMARY RESULTS \& FINDINGS}

The preliminary survey, conducted using multibeam \ac{SONAR}, successfully mapped the underwater vegetation in Maschsee Lake. Data processing was performed with Auto Clean (BeamworX) software. The resulting multibeam bathymetry \ac{SONAR} data revealed significant weed growth throughout the lake.

\begin{figure}[thpb]
  \centering
  \includegraphics[width=0.8\columnwidth]{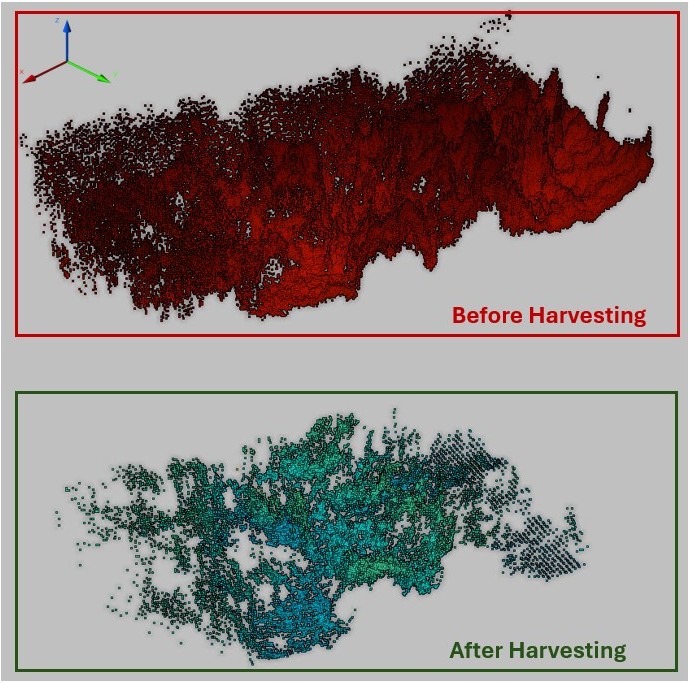}
  \caption{3D view of \ac{SONAR} point cloud data from an inspection area, showing the height difference in weed distribution detected by multibeam \ac{SONAR} \textbf{[Top]} before mowing and \textbf{[Bottom]} after mowing.}
  \label{fig:before_after_mowing}
\end{figure}

In an experiment, a designated area was first scanned with the \ac{USV} equipped with the \ac{SONAR}, followed by harvesting the weeds with a boat. A subsequent scan by the \ac{USV} showed notable changes in the weed distribution. Figure \ref{fig:before_after_mowing} illustrates the weed distribution using multibeam \ac{SONAR} bathymetry data before and after the mowing process. The average height difference between pre- and post-harvest was found to be \SI{80}{cm}.

\begin{figure}[thpb]
  \centering
  \includegraphics[width=\columnwidth]{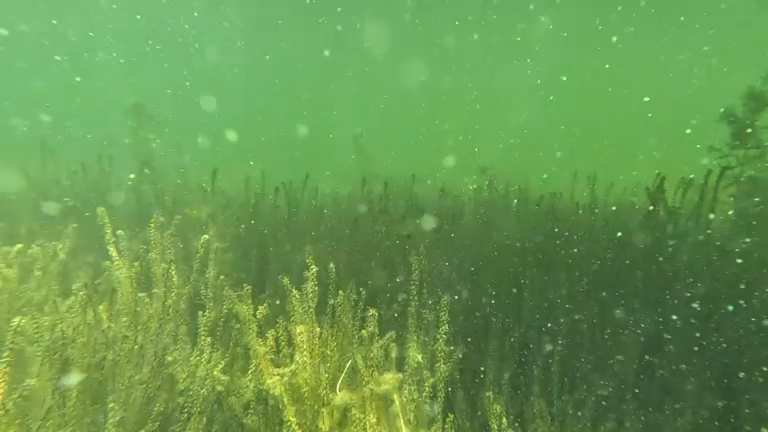}
  \caption{Underwater image showing weed at Maschsee Lake captured using GoPro camera.}
  \label{fig:gopro_maschsee}
\end{figure}

Underwater footage captured by a GoPro camera further validated the \ac{SONAR} findings for the same area. Figure \ref{fig:gopro_maschsee} displays these images, demonstrating the effectiveness of our system in detecting underwater vegetation and the benefits of the heterogeneous setup. The \ac{USV} collaborates with the operator  by providing a map that guides the mowing process, and it can estimate weed height, using the lake's depth map to determine the vegetation density.

\begin{figure}[thpb]
    \centering
    \subfigure[\label{fig:ladder_maschsee_gt}]{\adjincludegraphics[width=.49\columnwidth, trim={0 0 {.5\width} 0}, clip=true]{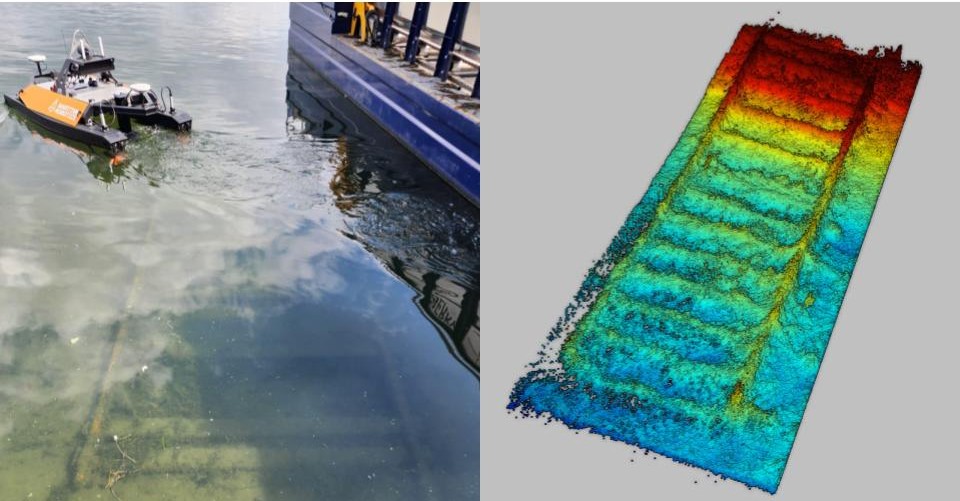}}
    \subfigure[\label{fig:ladder_beamworx}]{\adjincludegraphics[width=.49\columnwidth, trim={{.5\width} 0 0 0}, clip=true]{images/ladder_camera_sonar.jpg}}
    \subfigure[\label{fig:ladder_backscatter}]{\adjincludegraphics[width=.8\columnwidth, trim={0 0 0 0}, clip=true]{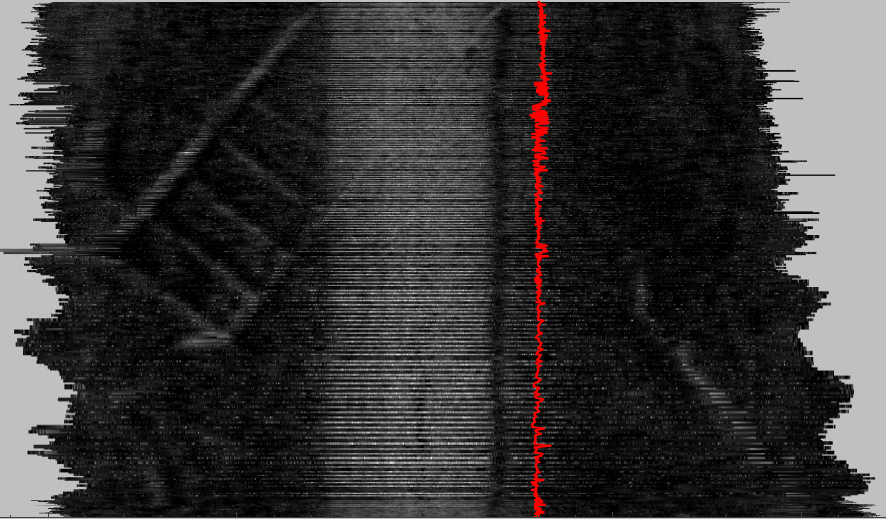}}
    \caption{Submerged Ladder [\textbf{a}] Camera image above water [\textbf{b}] Multibeam \ac{SONAR} 3D point cloud data [\textbf{c}] \ac{SONAR} Backscatter data.}
    \label{fig:ladder_maschsee}
\end{figure}

Furthermore, the multibeam \ac{SONAR} detected a submerged ladder, demonstrating the \ac{USV}'s capability to identify and guide operators away from hazardous or inaccessible areas. Figure \ref{fig:ladder_maschsee_gt} presents a camera image of the ladder, figure \ref{fig:ladder_beamworx} shows \ac{SONAR} 3D point cloud, and figure \ref{fig:ladder_backscatter} displays \ac{SONAR} backscatter data.

\section{FUTURE WORK}
Future work will focus on several key areas to enhance the weed harvesting process. One important aspect is the detailed analysis of backscatter data from the \ac{SONAR} to correlate it with weed volume and distribution across the lake. This analysis aims to refine our understanding of how \ac{SONAR} backscatter reflects weed density.

To validate these findings, we plan to incorporate additional data sources, such as satellite imagery, to detect shallow water bathymetry \cite{caballero2019retrieval}. Additionally, deploying a Remotely Operated Vehicle (ROV) equipped with an underwater camera and an acoustic positioning system will allow for precise location tracking of the ROV, providing ground truth for weed distribution.

Currently, weed detection relies on manual visual inspection by human operators, which is time-consuming. In the future, we will implement Artificial Intelligence (AI) techniques to automate weed detection. Techniques such as image segmentation, as demonstrated in \cite{garone2023seabed}, will be explored to enhance detection efficiency and accuracy.

Another critical area for future development is designing and implementing a path-planning algorithm to optimise the weed harvesting process. This algorithm will calculate the most efficient route for the boat, assisting the skipper in navigating through the weed clusters and harvesting them efficiently. It will also ensure that the conveyor belt capacity (\SI{15}{\metre\cubed}) is not exceeded, allowing the boat to reach the unloading station at the optimal time. Additionally integrating dynamic updates based on real-time data from the \ac{USV} will enable the algorithm to adjust the route in response to newly detected weed clusters during the mission. Developing this functionality will significantly enhance the efficiency and adaptability of the harvesting process.

\addtolength{\textheight}{-12cm}   




\section*{ACKNOWLEDGMENT}
This work is done within the HAI-x project, which is funded by the BMBF (Funding number: 01IW23003).


\bibliographystyle{ieeetr}
\bibliography{ref}


\end{document}